\documentclass[a4paper, 10pt, conference]{ieeeconf}      

\IEEEoverridecommandlockouts                              
\usepackage{graphics} 
\usepackage{graphicx}
\usepackage{amssymb}  
\usepackage{color}
\usepackage{listings}
\usepackage{multicol}
\usepackage{url}

\title{\LARGE \bf
Towards Python-based Domain-specific Languages for Self-reconfigurable Modular Robotics Research
}

\author{ \parbox{3.4 in}{\centering Mikael Moghadam and David Johan Christensen\\{\small Center for Playware\\Technical University of Denmark\\}{\tt\small \{mikm,djchr\}@elektro.dtu.dk}}
         \parbox{3.4 in}{\centering David Brandt and Ulrik Pagh Schultz\\{\small Modular Robotics Lab\\University of Southern Denmark\\}{\tt\small \{david.brandt,ups\}@mmmi.sdu.dk}}
}

\begin{document}

\maketitle
\thispagestyle{empty}
\pagestyle{empty}

\begin{abstract}

This paper explores the role of operating system and high-level languages in the development of software and domain-specific languages (DSLs) for self-reconfigurable robotics. We review some of the current trends in self-reconfigurable robotics and describe the development of a software system for ATRON II which utilizes Linux and Python to significantly improve software abstraction and portability while providing some basic features which could prove useful when using Python, either stand-alone or via a DSL, on a self-reconfigurable robot system.  These features include transparent socket communication, module identification, easy software transfer and reliable module-to-module communication. The end result is a software platform for modular robots that where appropriate builds on existing work in operating systems, virtual machines, middleware and high-level languages.

\end{abstract}

\section{INTRODUCTION}
Self-reconfigurable modular robotics (SMR) is an area of robotics concerning the design and creation of robotic modules that are able to communicate and cooperate in order to solve a given task  \cite{stoy} \cite{murata}. 

There are several fundamental challenges when dealing with SMR which seems to complicate software development. For instance, such a challenge is fully utilizing the computational resource of a robot configuration, coordinating module movement/actuation and developing software that deals with the parallel nature of an SMR robot on an (often) slow network connection (e.g., multihop IR). Also, one often has to deal with different types of hardware and mechanical designs given that research laboratories can have several different SMR platforms.
 \begin{table}[htb]
        \resizebox{8.5cm}{!}{
        \begin{tabular}{| c | c | c | c | c |}
        \hline
        System name and date    & Main processing unit  & Clock [MHz]   & Flash [KB]    & RAM [KB]      \\
        \hline
        ATRON I, 2004           & Atmel ATMEGA128       & 16            & 128           & 4             \\
        GZ-I, 2006              & P89LPC920             & 18            & 8             & 0.256         \\
        SUPERBOT, 2006          & Atmel ATMEGA128       & 16            & 128           & 4             \\
        M-TRAN III, 2007        & Renesas HD64F7047     & 50            & 256           & 12            \\
        Molecubes, 2007         & Atmel ATMEGA16        & 16            & 16            & 1             \\
        CKBot, 2008             & Microchip PIC18F2680  & 40            & 64            & 3.3           \\
        RoomBots, 2009          & Microchip PIC33FJ128  & 7 (40 MIPS)   & 128           & 16            \\
        Odin, 2009              & Atmel AT91SAM7S256    & 55            & 256           & 64            \\
        ATRON II, 2009          & Microblaze (32bit)    & 100           & 8192          & 65535         \\
        \hline
        \end{tabular}
        }
        \caption{Resources available on several recent SMR robots \cite{mirko}}
        \label{modular:table:capabilities}
\end{table}

Software development for SMR systems is not made easier by the fact that until very recently, as illustrated by Table \ref{modular:table:capabilities}, modules have been very limited in terms of computational resources. Not because more powerful hardware is unavailable or impossible to utilize, but likely because priority has been given to the cost effectiveness and robustness of the modules and a desire to keep the hardware complexity at a minimum. 

The higher the levels of abstraction we are working with are, the easier it usually gets to deal with the complexity of the algorithm/task at hand \cite{miller}. An increasingly popular way of trying to introduce abstraction tailored to the challenges of SMR, is by creating (embedded) Domain Specific Languages \cite{mirko} \cite{simpar} \cite{meld} \cite{ldp}.  Concretely, for the ATRON robot we have previously developed the DynaRole domain-specific language (DSL) for modular robots, which is a high-level role-based language that compiles to a space-efficient bytecode format executed by an dedicated virtual machine \cite{fastflexible}.

\begin{figure}[htp!]
        \centering
        \includegraphics[scale=0.13]{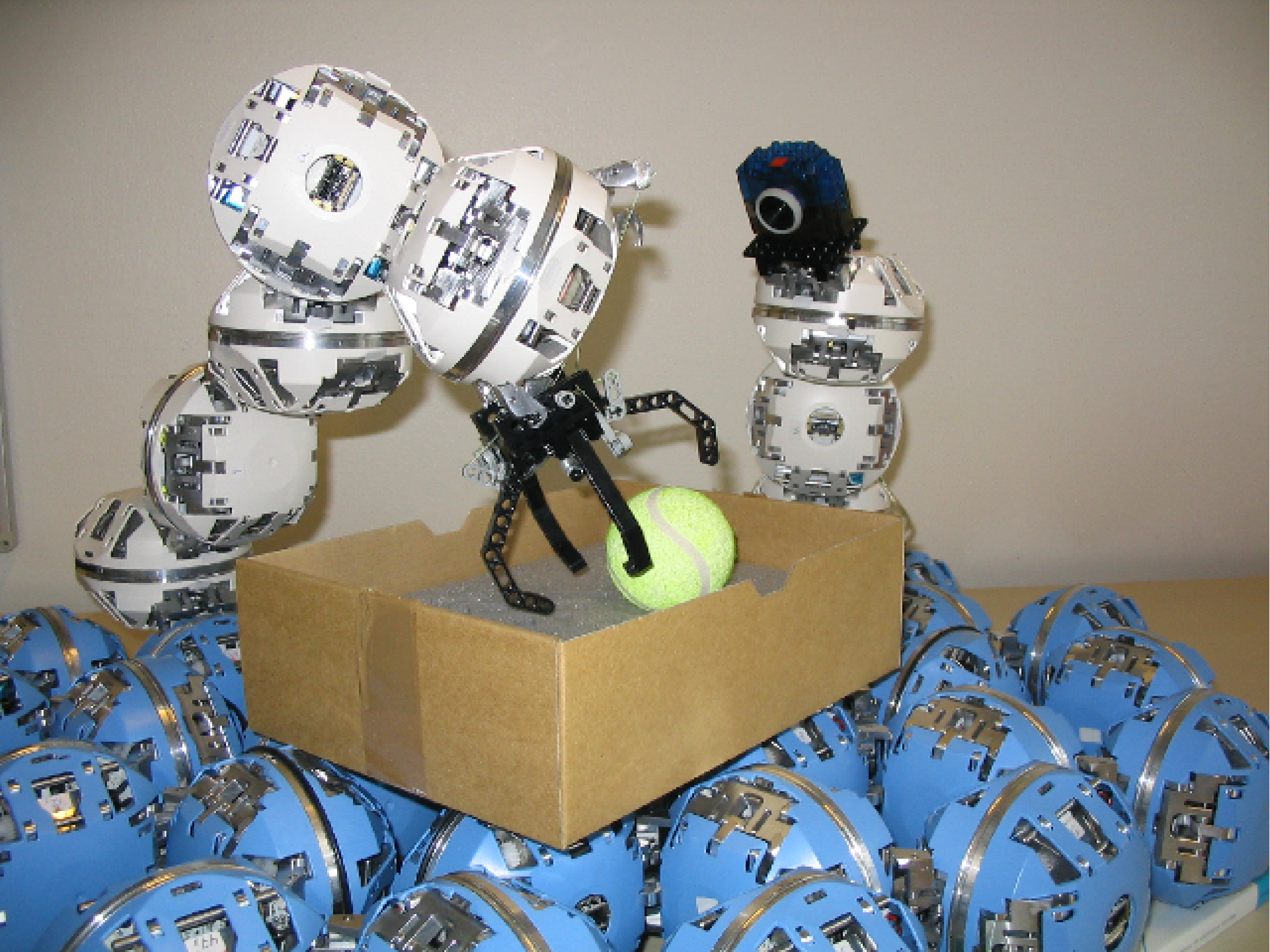}
        \caption{Arms and conveyor belt constructed of ATRON modules \cite{davidphd}}
        \label{fig:atronrobot}
\end{figure}

The latest version of the ATRON II modular robot developed at the University of Southern Denmark (see Figure \ref{fig:atronrobot}), has taken a great leap in terms of its computational hardware platform \cite{brandt} \cite{microblaze}. This means that unlike previous and other modular robots, we have the option of utilizing many of the amenities common in less constrained areas of robotics and general computing. We have therefore chosen to explore another route than the traditional DSLs, which are compiled to machine code or designed for very constrained embedded virtual machines \cite{mirko} \cite{controldiffusion} \cite{meld}, namely, to use an operating system that will enable us to utilize a modern full-fledged interpreted language. That is the topic of this paper, in which we equip the ATRON II modules with Linux and Python, in order to experiment with high-level SMR software development.\footnote{This paper is based on the first author's MS thesis. We refer to this document for additional details~\cite{mikael-ms}.}  Specifically, we create a high-level framework which encapsulates some of the details of the ATRON II system and which can be used later on as a foundation for creating DynaRole-like abstractions and semantics (roles and behaviours), effectively bypassing some of the limitations that DynaRole has due to the specificness of its virtual machine. 

\section{Software frameworks for the ATRON}

\subsection{Domain Specific Languages for SMR}
\label{dslformodularrobots}
In order to get a framework more geared towards SMR development, a Domain Specific Language (DSL) was developed by Bordignon et al. \cite{mirko} by the name of DynaRole.

Bordignon et al. also implemented a virtual machine intended for use on ATRON I modules, by the name of \emph{Embedded-Distributed Control Diffusion Virtual Machine} (E-DCD VM) \cite{controldiffusion} \cite{fastflexible}. The introduction of a virtual machine allows the DSL to compile to byte-level code, meaning that portability can be obtained by porting the virtual machine to other platforms, in much the same way portability is obtained in Desktop software development (e.g., JVM, Python, .NET CLR etc.). This furthermore has the benefit that once a single module is programmed, the bytecode can be automatically transferred to other modules in an easy fashion through the module-to-module communication channels. In other words, the entire robot configuration can be programmed from a single point rather than having to program each module individually. Single point access/programming in itself does not require a virtual machine, since it could also be achieved by using a boot-loader and then distributing a new binary image to be booted, or even by distributing new 'ordinary' compiled executables. The bootloader method requires that the module be rebooted, and transferring a new executable would mean that the entire program would have to be transferred even when performing minute changes (unless one implements a form of bin-diff and merge), while only updated bytecode fragments need to be transferred with the virtual machine approach. Moreover, tests performed show that rebooting the module incurs a significant overhead (three-fold) in terms of time spent, which is an important factor, given that developing software for modular robots is often an incremental and iterative process. 

The virtual machine also enables the appending of meta-information to bytecode, which is utilized by the software to gain information about the configuration when bytecode emanates through it. Information about the origin is retained so that modules are able to share a common coordinate system and have general information about the structure of the configuration. 
By utilizing the DynaRole DSL one can write programs in a higher-level language (rather than bytecode) where \emph{roles} and \emph{behaviours} are the primary form of abstraction. Roles are assigned to the modules in the structure. \emph{Behaviours} are associated with roles. For instance a \emph{Move} behaviour can be associated with the \emph{Wheel} role. 
Due to a limited set of primitives, complex computations have to be performed by `external' code (C code). A role is similar to a class in the sense that it can declare a number of member functions and can inherit members from parent roles. Roles are stateless.

Other examples of DSLs for modular robotics are Meld \cite{meld} and LDP \cite{ldp}.  Although designed for execution on modules with severe resource constraints, they currently run in simulation only, and there are no specific considerations on how to implement the language runtime for these languages.

\label{TinyOS}
\subsection{TinyOS}
The previous ATRON hardware generation (``ATRON I'') used the TinyOS~\cite{tinyos} operating system as a means of hardware abstraction and this operating system was ported to ATRON II in order to both retain the expertise accumulated by the researchers and the SMR software created \cite{mirko}~\cite{simpar}.

TinyOS was designed for use in wireless sensor networks. Sensor networks are usually wireless ad-hoc networks implementing a \emph{multihop} routing algorithm for package routing between sensors in the network. In this sense, a robot built of ATRON modules can essentially be seen as a sensor network, since sensor communication in such a robot is both wireless and ad-hoc utilizing multihop architecture. TinyOS would perhaps seem to be an appropriate choice because it is specifically designed for use in such systems. However being tailored to run on limited embedded platforms, for which the primary purposes are to essentially collect data, does inflict certain distinct limitations on the operating system. It should be noted that ATRON I only had 4Kb of RAM. This means that obtaining a high level of concurrency by way of software multi-threading would be difficult due to stack size limitations. However, TinyOS solves this by offering stackless threads and a flat memory model with no kernel space.

TinyOS is written in nesC which is a variation of the ANSI C language created for TinyOS and tailored for use in memory limited sensor networks. 
The level of network transmission that TinyOS (version 2.x) provides is \emph{best effort} and multihop. 
No guarantees are given in terms of receiver delivery and no maximum routing time or hop count can be given. 
TinyOS must be statically linked with the software that it is intended to run, meaning that it is not an operating system that is running continuously on a platform in which a piece of software is then loaded and executed explicitly.

When designing ATRON II, the move to an FGPA/Microblaze platform meant that the existing software and software development experience would in large parts be lost, unless TinyOS were ported to the Microblaze platform. The choice was therefore made to  port it to Microblaze. Later, TinyOS was ported to the ARM platform, enabling its use on ARM based SMR modules of the Modular Robotics Lab. Porting TinyOS to ARM took approximately 4 months full-time work for one university professor.
TinyOS provides a fundamental level of hardware abstraction, and some software abstraction (nesC vs. plain C). Portability is basically attained by porting TinyOS to various platforms, enabling existing software to be used on them.\footnote{We are currently also experimenting with an alternative approach to portability and reusability of software for resource-constrained modular robots, based on the reuse portable library \emph{Assemble-and-Animate} developed by Christensen et al.~\cite{ase}. This library has been developed with portability and reusability in mind and contains many algorithms and tools useful for SMR software development.}

\section{From virtual machine to operating system}

The software stack consisting of TinyOS, E-DCD VM and DynaRole serve to provide abstraction, portability and reusability. TinyOS provides a basic level of hardware abstraction and was a choice made primarily with the tight resource constraints of ATRON I in mind. 
TinyOS has a somewhat limited further potential for the ATRON platform, which in hindsight is unsurprising seeing as how it is designed for sensor networks. Consider for instance this fact: \emph{"TinyOS doesn't work well with long, uninterrupted pieces of code, CPU-intensive applications can cause problems."}~\cite{tinyos-limitations}. There should be no need for imposing such limitation on a hardware platform having a 100MHz SMP processor and 64MB RAM.
The E-DCD VM is a modern step in the direction of increased abstraction and portability, but it suffers somewhat from the simplicity of the VM and the DSL compared to Java JVM and Python. 
Due in part to the required implementation effort and in part to the severe memory restrictions of the ATRON~I platform, the DynaRole language lacked basic constructs such as local variables, loops, and functions.
Moreover, the E-DCD~VM execution model essentially provides a global environment in which mobile bytecode fragments can be installed and modified; this approach is highly flexible but also somewhat fragile and 
makes it difficult to implement more complex scenarios where multiple behaviors are running on the same set of modules.
Given these experiences and the more powerful hardware of the ATRON~II modules, we now define a revised set of requirements for an execution platform for modular robots.

\subsection{Requirements}

We see the ATRON~II robot as a platform for research in SMR, both for new hardware concepts (this is supported by the FPGA), for new control algorithms, and for new programming languages supporting these control algorithms.  This usage scenario gives rise to the following requirements on the software platform.
\begin{itemize}
\item{{\bf Software development support} A development environment in which it is easy to develop new control software for the robot, in the language preferred by the developer.}
\item{{\bf Multiple language/VM support} There exists many virtual machines and languages, for DSL development it would be useful to have the ability to experiment with them and be able to choose the language or virtual machine that is best for the task at hand.}
\item{{\bf Process abstraction} We want applications to be shielded from each other, so that one faulty application does not interfere with other applications, or bring down the entire system.  This includes the ability to cleanly take down a process and start a new one.}
\item{{\bf SMP support} The ATRON~II now has an FPGA at its heart enabling the use of multiple Microblaze cores for speed or robustness, so the use for multiple CPUs must be supported.}
\item{{\bf Hardware support} The ATRON~II has been designed for easy expansion with new hardware, if we connect existing hardware to a module it is preferable that the development effort to support this new hardware is minimized.}
\end{itemize}
These requirements obviously go beyond what is offered by standard embedded operating system and virtual machines, but are a perfect match for an operating system such as Linux.

\subsection{Solution}
In order to meet the requirements just outlined, we choose to run Linux on the ATRON robots, since that will potentially support them all: software development, multiple languages and virtual machines, processes, SMP and standard hardware is all supported very well by Linux.
In order to experiment with high-level DSL development, we initially choose to utilize the Python interpreter on the platform, because we believe it to be a modern multi-paradigm dynamic and high-level language which could potentially provide many benefits in the creation of DSLs.
The approach we are taking is similar to that of many other modern robotic frameworks, namely the so called middleware approach, in which an operating system is required to provide the fundamental means of abstraction against which the framework is then designed \cite{orocos} \cite{player} \cite{extendedrt}. 

Linux adopted the Microblaze architecture in version 2.36.1. The ability to use Linux and Python is however not possible just because of the Microblaze architecture, but also because of the increased RAM and flash. The RAM benefits are especially significant, given that Linux Kernel requires at least 8 MB RAM to run.
Linux is a standards compliant operating system which brings with it numerous benefits that are immediately useful in the context of almost any robotics application: 

\subsubsection{Task management}
Linux provides a clear separation between the kernel and user applications which means that a faulty user application can not access kernel data structures.

\subsubsection{Uniform interface to resources}
Most resources can be accessed through file descriptors/handles, no matter how the functionality is implemented in hardware. One of the benefits of this is that target environments can be easily simulated, because the interface on the target system will be the same (or easily reproducible) on the development system.

\subsubsection{Resource arbitration in case of contention}
Linux comes with a variety of scheduler options for task and I/O scheduling.

\subsubsection{Peripheral support}
Driver development is done by using the kernel API, meaning that once the driver is developed for one platform (ARM for instance) it usually works on all other supported platforms as well, requiring little or no extra development work.

\subsubsection{Software Ecosystem}
Linux is more than a kernel. It is a large library of software. One great thing about the portability of Linux is that the software that runs on Desktop Linux most likely also runs on Embedded versions of Linux.

\subsubsection{Real-time support}
Linux offers low-latency predictable task scheduling, which, in the context of robotics, can be an important feature. As a stock kernel, Linux offers near Real-Time support (ISR cannot be interrupted), but there exists a patch-set which will make the kernel fully preemptible.

\section{The role of the virtual machine}
Python provides a much more powerful and complete virtual machine/interpreter than the E-DCD~VM with DynaRole, yet it lacks some of its SMR features such as: the ability for the application to adapt to the state of the module itself and the immediately surrounding modules (role adaptation), automatic transfer of updated applications throughout the robot configuration, unique identifiers for each module in the configuration, direct communication with applications on other modules, and broadcast of data to all applications on other modules. To remedy this shortcoming, we have developed a Python framework called \emph{Moduleservice}. Moduleservice runs as a process on each module, with which other applications can communicate in order to obtain information about the module itself and other modules in the configuration. Moduleservice is not a complete replica of the E-DCD VM features in that it lacks some of its functionality, but it provides some features that the E-DCD VM does not.
Moduleservice is meant to augment the Python interpreter with some features targeting SMR, which a DSL targeting the Python interpreter could then utilize. One could for instance port DynaRole to the Python Moduleservice, which would give a massive increase in the expressive power of the language.

\subsection{Moduleservice features}

The Moduleservice framework provides the following features:

\setcounter{subsubsection}{0}
\subsubsection{A socket for communication}
Moduleservice acts as a socket-server to which other applications running on the module can connect. Having connected, an application sends specific textual commands over the socket in order to communicate with the Moduleservice application (and by extension, other modules). The Moduleservice application responds to commands received by issuing textual commands over the same socket.

\subsubsection{Module state information}
Module state in this case means the state of the physical module, e.g.,\ whether connectors/grippers are opened or closed. Applications can ask for the state of the module on which they are currently running, but also for the states of neighbouring modules.

\subsubsection{Automatic update of the Moduleservice application to neighbouring modules that have an older version}
If a neighbouring module has an older version of Moduleservice, it gets sent the updated version. 

\subsubsection{Unique identification of individual modules}
In the process of transmitting a new version of Moduleservice to neighbouring modules, the identifier of the module contained within Moduleservice gets modified so that the receiving module will have a unique identifier.

\subsubsection{Communicate with individual applications on neighbouring modules}
An application running on a module must register itself with Moduleservice upon connection. In this way Moduleservice will have knowledge of which applications are running on the module. Applications can address data directly to specific applications on neighbouring modules.

\subsubsection{Broadcast data to applications running on neighbouring modules}
Applications can ask the Moduleservice running on neighbouring modules to broadcast data to all running applications.

\subsubsection{Transfer (text) files}
An application can ask Moduleservice to transfer a text file (such as a Python code file) from the current module to a neighbouring module

\subsubsection{Execute code on neighbouring modules}
An application can ask a neighbouring modules' Moduleservice to execute a line of Python code supplied by the application.

\subsubsection{Start applications on neighbouring modules}
An application can ask a neighbouring modules' Moduleservice to start a specific Python application. For instance, a newly transferred Python source file.

\subsection{Implementation}
The implementation of Moduleservice is based upon two different communication protocols: one for communicating with Moduleservices on neighbouring modules and one for communicating with applications running on the module (sockets). There are two because the requirements are different and the formatting and layouts of the protocols are different.

The module-to-module communication handling is straightforward. Each IR port has its own protocol handler running in its own thread, which parses incoming messages, verifies the checksum, schedules a potential acknowledgement message and then goes on to handle the actual message. Each protocol handler maintains a transmit buffer. When a message is sent, the protocol handler waits for an acknowledgement message. If it arrives within a specified interval, the message gets removed from the transmit buffer. If not, it is retransmitted. 
The use of Python's advanced string handling routines, socket support, easy file-reading and writing, multi-threading and object-oriented model significantly facilitated the implementation compared to an implementation in C. 
We are also making heavy use of Python's standard data structures such as lists, tuples and dictionaries (powerful dynamic arrays, immutable arrays and hash-tables).

Moduleservice does not provide module orientation information or allow just the specific bytes of an application to be modified easily. It also does not enable applications to broadcast data to modules having specific roles, as the role abstraction model of the \emph{DynaRole} language is not yet implemented. 
Moduleservice provides reliable communication, provides transfer of files, and allows applications to talk to each other by name, none of which the E-DCD VM provides. Moduleservice has a language-agnostic interface. 

\section{High-level language proposal}

\begin{figure*}
\begin{minipage}[t]{0.45\textwidth}
\footnotesize
\begin{verbatim}
role Head extends Module {
 require (self.center == $NORTH_SOUTH);
 startup initialize(_) {
  handle $EVENT_HANDLER_1 $EVENT_HANDLER_3 {
   Wheel.evade(0); 
   (self.sleepcs(25));
  };
  (self.enable($EVENT_HANDLER_1));
  (self.enable($EVENT_HANDLER_3));
 }
}

abstract role Wheel extends Module { 
 abstract constant connected_dir;
 abstract constant turn_dir;
 abstract constant evasion_dir;
 require (self.center == $EAST_WEST);
 require (sizeof(self.connected(
          connected_dir)) == 1);
 behavior move(_) {
  self.$TURN_CONTINUOUSLY(turn_dir);
 }
 command evade(_) {
  self.$TURN_CONTINUOUSLY(evasion_dir);
  (self.sleepcs(25));
 }
}

role RightWheel extends Wheel { 
  turn_dir = 150; evasion_dir = -100;
  connected_dir = $EAST;
}

role LeftWheel extends Wheel {
  turn_dir = -150; evasion_dir = 100;
  connected_dir = $WEST;
}
\end{verbatim}
\end{minipage}
\begin{minipage}[t]{0.49\textwidth}
\footnotesize
\begin{verbatim}
class Head(Module):
    @require
    def dir(self):
        return self.center == NORTH_SOUTH
    @handle([PROXIMITY_1,PROXIMITY_3])
    def proximity(self):
        Wheel.evade()
        self.sleepcs(25)

class Wheel(Module):
    @require
    def dir(self):
        return self.center == EAST_WEST
    @require
    def oneConnection(self):
        return len(self.connected(connected_dir)) == 1
    @behavior
    def move():
        self.TURN_CONTINUOUSLY(turn_dir)
    @command
    def evade():
        self.TURN_CONTINUOUSLY(evasion_dir)
        self.sleepcs(25)

class RightWheel(Wheel):
    def __init__(self):
        self.turn_dir = 150
        self.evasion_dir = -100
        self.connected_dir = EAST

class LeftWheel(Wheel):
    def __init__(self):
        turn_dir = -150
        evasion_dir = 100
        connected_dir = WEST
\end{verbatim}
\end{minipage}
\caption{DynaRole programming: original syntax (left) and proposed embedded python syntax (right)}
\label{fig:dynarole-variants}
\end{figure*}

A key motivation for this work was to enable the development of
software for modular robots using embedded domain-specific languages.
With Linux ported to the ATRON and Python running with the
ModuleService framework, an embedded DSL for DynaRole can be designed
and implemented.  Programs written in this DSL would be distributed
and updated using ModuleService and could communicate using the
ModuleService socket abstraction. 
We are currently investigating how the implementation can be done in the most effective way.

As a concrete example, see Figure~\ref{fig:dynarole-variants} which shows a
DynaRole program for obstacle evasion written using the previously developed, dedicated
bytecode compiler (left-hand side) and using a hypothetical Python
embedded DSL syntax (right-hand side).  This program implements
obstacle evasion for a 3-module car-like robot: One module plays the
role of ``Head'' which is at the front and uses proximity sensors to
detect obstacles, when an obstacle is detected it triggers the
``evade'' methods in all wheel modules.  Two modules play the role of
``Wheel'' which by default performs the ``move'' behavior where the
module is rotating, pushing the robot forward like a wheel.  These modules can also
perform the ``evade'' behavior which causes them to reverse their
direction temporarily (methods are mutually exclusive).  The abstract
``Wheel'' role has two subroles that define the concrete behavior
using constants.  The roles are dynamically assigned based on the
invariants in the ``require'' clauses of each role, the role
abstraction allows this same program to be used on several robots with
similar morphology.

Although this embedded DSL has not been implemented, we can perform an initial analysis of what its properties would be.  From the point of
view of efficiency the Python interpreter is not likely to be
significantly faster than the highly-simplified DynaRole bytecode
interpreter, but speed is not an issue for the relatively slow-moving
ATRON modules.  Program size is another matter, it is critical since
programs must be communicated over infrared which can be very slow.  We
observe that the bytecode-compiled version of the program shown in
Figure~\ref{fig:dynarole-variants} (left-hand side) takes up 156 bytes
in total (in E-DCD VM these would be broken up into self-contained
fragments before transmission), whereas the proposed equivalent Python
program (right-hand side of the figure) could be transmitted in
gzipped format, which would take up 350 bytes.  Although by no means a
conclusive study of program sizes, we believe these numbers to be
representative: transmitting compressed source code will have a
significant overhead but will still be feasible.  Last, from a
language design perspective, although the Python-based DSL is fairly
concise there is a significant amount of syntactic noise compared to
the original DynaRole program.  This seems to be a typical problem
with Python-based DSLs, for which reason we are currently
experimenting with other high-level languages such as Ruby.

\section{Conclusion and Future Work}
The purpose of this paper has been to investigate the use of Python as a way of obtaining software abstraction, portability and reuse in self-reconfigurable modular robotics and comparing it with the blossoming trend of stand-alone DSLs in modular robots, both as an alternative but also as a supplement. We did so by installing Linux on the ATRON II modules developed by the Modular Robotics Lab at the University of Southern Denmark. 
Doing so, the interactiveness of the ATRON II module with Linux has risen to a new level. The developers can open a terminal interface to the module and, through it, create software or do debugging directly and interactively on the modules themselves. They have a whole plethora of existing software and applications that they can utilize in their research and development, and they can even do future software development in a very high-level programming language, which no doubt will make the software more versatile, portable and reusable.
The Moduleservice framework, while being `only' 451 lines of python code, already has some fairly complex features, such as encapsulation of module-to-module communication, allowing applications on separate modules to communicate via sockets, and the ability to transfer source code from one module to another. The key insight that we take for granted on many other computing platforms, but that is often ignored on modular robots, is that we did not first have to develop a language or a virtual machine, but rather relied on a powerful operating system and an existing modern virtual machine to do the ``heavy lifting.'' The practical consequence of this is that we are able to enjoy a lot of the power and the abstractions of Linux and Python without having implemented them explicitly beforehand.

The lesson learned is that both the host language {\em and} its operating system provide benefits for the embedded DSL.  Some of the features that we require for our embedded DSL for modular robot programming can be provided by a language such as Python, but a significant number of features are also provided by the underlying operating system.  The more flexible and powerful the language and operating system, the easier it becomes to implement the DSL abstractions required for the specific domain.

In terms of future work, implementing the E-DCD VM features that are still missing from Moduleservice is an evident future improvement, especially the DynaRole role-based programming model could likely be a beneficial addition implemented as a DSL in Python atop Moduleservice. 
The module-to-module communication bandwidth utilization has had little priority during development, and is also an obvious area of potential improvement. 

\subsubsection*{Acknowledgements} We would like to thank the DSLRob reviewers for their insightful and constructive comments which were very helpful in improving our paper.


\end{document}